\documentclass[manuscript,screen]{acmart}

\usepackage{subcaption}

\usepackage{algorithm}
\usepackage[noend]{algpseudocode}

\usepackage[final]{pdfpages}
\AtBeginDocument{%
  \providecommand\BibTeX{{%
    \normalfont B\kern-0.5em{\scshape i\kern-0.25em b}\kern-0.8em\TeX}}}

\setcopyright{rightsretained}
\begin{document}
\copyrightyear{2021}
\acmYear{2021}
\acmConference[COMPASS '21]{ACM SIGCAS Conference on Computing and Sustainable Societies (COMPASS)}{June 28-July 2, 2021}{Virtual Event, Australia}
\acmBooktitle{ACM SIGCAS Conference on Computing and Sustainable Societies (COMPASS) (COMPASS '21), June 28-July 2, 2021, Virtual Event, Australia}\acmDOI{10.1145/3460112.3471947}
\acmISBN{978-1-4503-8453-7/21/06}
 
\title{ElephantBook: A Semi-Automated Human-in-the-Loop System for Elephant Re-Identification}

\author{Peter Kulits}
\email{kulits@caltech.edu}
\affiliation{%
  \institution{Caltech}
  \country{N/A}
}

\author{Jake Wall}
\email{jake@maraelephantproject.org}
\affiliation{%
  \institution{Mara Elephant Project}
  \country{N/A}
}

\author{Anka Bedetti}
\email{anka@elephantsalive.org}
\affiliation{%
  \institution{Elephants Alive}
  \country{N/A}
}

\author{Michelle Henley}
\email{michelephant@savetheelephants.org}
\affiliation{%
  \institution{Elephants Alive}
  \country{N/A}
}

\author{Sara Beery}
\email{sbeery@caltech.edu}
\affiliation{%
  \institution{Caltech}
  \country{N/A}
}

\renewcommand{\shortauthors}{Kulits, Wall, Bedetti, Henley, \& Beery}

\begin{abstract}
African elephants are vital to their ecosystems, but their populations are threatened by a rise in human-elephant conflict and poaching. Monitoring population dynamics is essential in conservation efforts; however, tracking elephants is a difficult task, usually relying on the invasive and sometimes dangerous placement of GPS collars. Although there have been many recent successes in the use of computer vision techniques for automated identification of other species, identification of elephants is extremely difficult and typically requires expertise as well as familiarity with elephants in the population. We have built and deployed a web-based platform and database for human-in-the-loop re-identification of elephants combining manual attribute labeling and state-of-the-art computer vision algorithms, known as ElephantBook. Our system is currently in use at the Mara Elephant Project, helping monitor the protected and at-risk population of elephants in the Greater Maasai Mara ecosystem. ElephantBook makes elephant re-identification usable by non-experts and scalable for use by multiple conservation NGOs.
\end{abstract}



\begin{teaserfigure}
  \includegraphics[width=\textwidth]{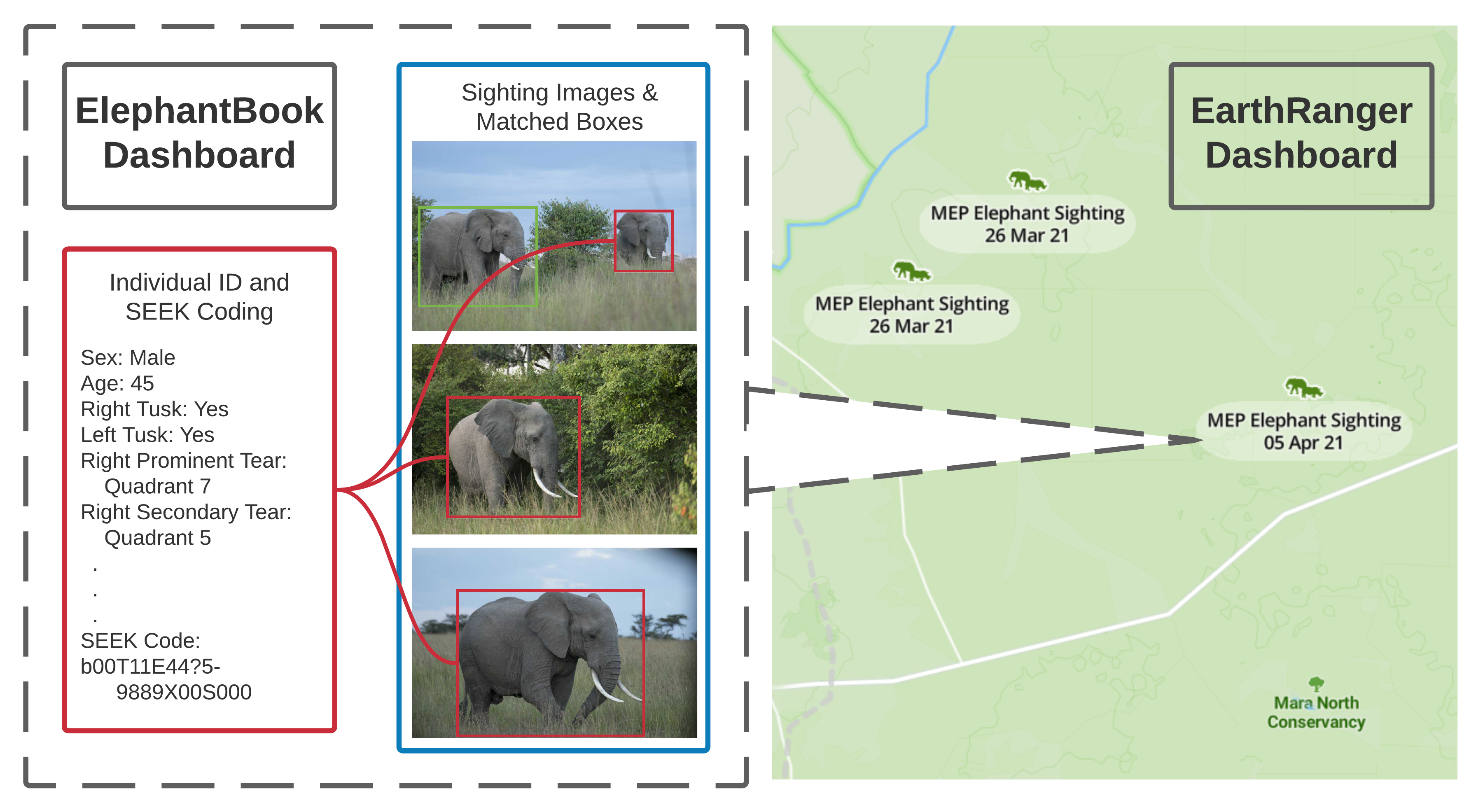}
  \caption{ElephantBook: a system for human-in-the-loop elephant re-identification. Our system can be linked to the EarthRanger conservation land management platform \cite{earthranger}, and it helps humans efficiently monitor elephant populations and locations from elephant sightings in the wild.}
  \label{fig:teaser}
\end{teaserfigure}
\maketitle

\section{Introduction}
Reliable wildlife population monitoring is critical for effective conservation and species management. Accurate measurement of wildlife density and distribution across landscapes provides insight into trends and ecological processes such as population growth, fecundity, survival, mortality, and density-dependent regulation. A range of measurement techniques have been developed which include aerial surveys, camera trap networks, ground survey techniques, and individual-based re-identification (e.g., spatially explicit mark-recapture \cite{royle_spatial_2013}). Individual-based recognition techniques can also be used in behavioral studies and human-wildlife conflict cases. The emergence of computational systems based on image algorithms has recently made traction enabling re-identification of certain species (e.g., whales, sharks, zebras, seals, lynx, and sea turtles) that present distinct morphology or patterns (e.g., contours, spots, or stripes) that facilitate visual separability among individuals  \cite{berger-wolf_wildbook_2017}. However, many species are cryptic and difficult to observe, difficult even for experts to distinguish, or currently lack sufficient training data for application of computer-vision approaches. 

Vital to their ecosystems, African elephants are especially important to monitor closely; they are considered ecosystem engineers who have the capacity to shape the environments in which they live, and their population density and distribution can impart multiple cascading effects on ecosystems, biodiversity, and tourism-based economies \cite{pringle2008elephants, haynes2012elephants, naidoo2016estimating}. Both species of African elephants are threatened: the savanna elephant (\emph{L. africana}) is endangered, and the forest elephant (\emph{L. cyclotis}) was recently listed as critically endangered by the IUCN Red List \cite{noauthor_iucn_nodate}. Some populations have suffered as much as 62\% population loss in recent years \cite{maisels_devastating_2013} with the ivory trade and associated poaching being the main drivers of their decline. Characterizing elephant population demographics across their range is therefore essential to conservation of the species.

Ecologists have recently attempted to create a general re-identification method that can be used by non-experts. The best known of these methods is System for Elephant Ear-pattern Knowledge (SEEK) coding, developed by Elephants Alive \cite{bedetti2020system}, which uses manual attribute labels such as sex and the presence/absence of tusks to improve the accuracy and efficiency of re-identification.
The Mara Elephant Project, in collaboration with the California Institute of Technology and Elephants Alive, has developed a semi-automated ensemble visual-recognition system using photographs taken by rangers and research field teams along with manual SEEK attribute labeling. ElephantBook is a novel online software solution with the goal of making elephant re-identification accessible by non-experts and scalable to multiple conservation NGOs. 

\section{Background}

\subsection{The Greater Mara Ecosystem}
The Greater Mara Ecosystem (GME) in Kenya is a critical ecosystem given its biodiversity, large wildlife populations, and rich cultural history. It forms the northern extent of the annual migration of 2.2 million wildebeest, zebra, and gazelle from the Serengeti, and it is the most-visited tourist destination in Kenya. The most recent census results estimate there are 2,493 elephants in the GME \cite{kenya_wildlife_service_aerial_nodate}. Elephants typically live in family units consisting of related females and their offspring. Adult male elephants roam alone or in bachelor herds after they've reached an age of sexual maturity. 
Despite its status as one of the most beautiful and important wildlife areas in the world, the GME faces significant conservation threats: 374 elephants have been illegally killed since 2012, and there has been a 60\% increase in recorded incidents of human-elephant conflict since 2017 (Mara Elephant Project unpublished data). The expansion of agriculture, infrastructure, and human populations is infringing into current elephant ranging areas and severing movement corridors. 50\% of elephant range now falls outside of protected areas \cite{wall_human_2021}.

\subsection{The Mara Elephant Project}
The Mara Elephant Project (MEP), established in 2011, protects savanna elephants and works to conserve the greater Maasai Mara ecosystem (GME) in Kenya. MEP, in conjunction with the Kenyan Government, has deployed ranger teams to follow the locations of elephant groups fitted with real-time GPS tracking collars, which has led to the arrest of 373 poachers, the seizure of 1,676.5 kg of ivory, and the identification of core movement patterns of approximately 500 elephants \cite{noauthor_mep_nodate}. MEP also frequently dispatches rangers to help mitigate conflicts involving ``crop-raiding'' elephants. Identifying which individual elephants are involved in crop-raiding is important because raiders are typically repeat offenders. Ongoing field monitoring, data analysis, and conservation efforts are needed to ensure the long-term survival of elephants and the overall GME.

\subsection{Elephants Alive}
Elephants Alive is a South Africa-based non-profit organization that operates across the Greater Limpopo Transfrontier Conservation Area and the southern part of Mozambique. Although officially registered with the Kruger National Park in 2003, Elephants Alive draws on data collected over a quarter of a century. Its work contributes to the long-term survival of African elephants through a greater understanding of the complex relationship between elephants and the ecosystems they occupy and by identifying science-based solutions that enable elephants and people to coexist.

\subsection{EarthRanger}

Vulcan’s EarthRanger \cite{earthranger} is a real-time system for conservation-related data aggregation, storage, visualization, and dissemination \cite{wall_novel_2014}. It includes tracking data from wildlife, rangers, and vehicles, and it records ``Events,'' which range from human-elephant conflict to poaching to illegal logging. Events are reported from the field using a mobile application called Cybertracker; these reports include the time, the location, and information specific to each event type. The Mara Elephant Project, along with many other NGOs in Sub-Saharan Africa, now uses EarthRanger daily to record elephant Group Sightings, including information about group size and composition. However, EarthRanger does not currently support or have any type of interface for individual-based elephant re-identification.

\begin{figure}[ht]
\begin{center}
\begin{tabular}{c} 
\includegraphics[height=13.3cm,angle=90]{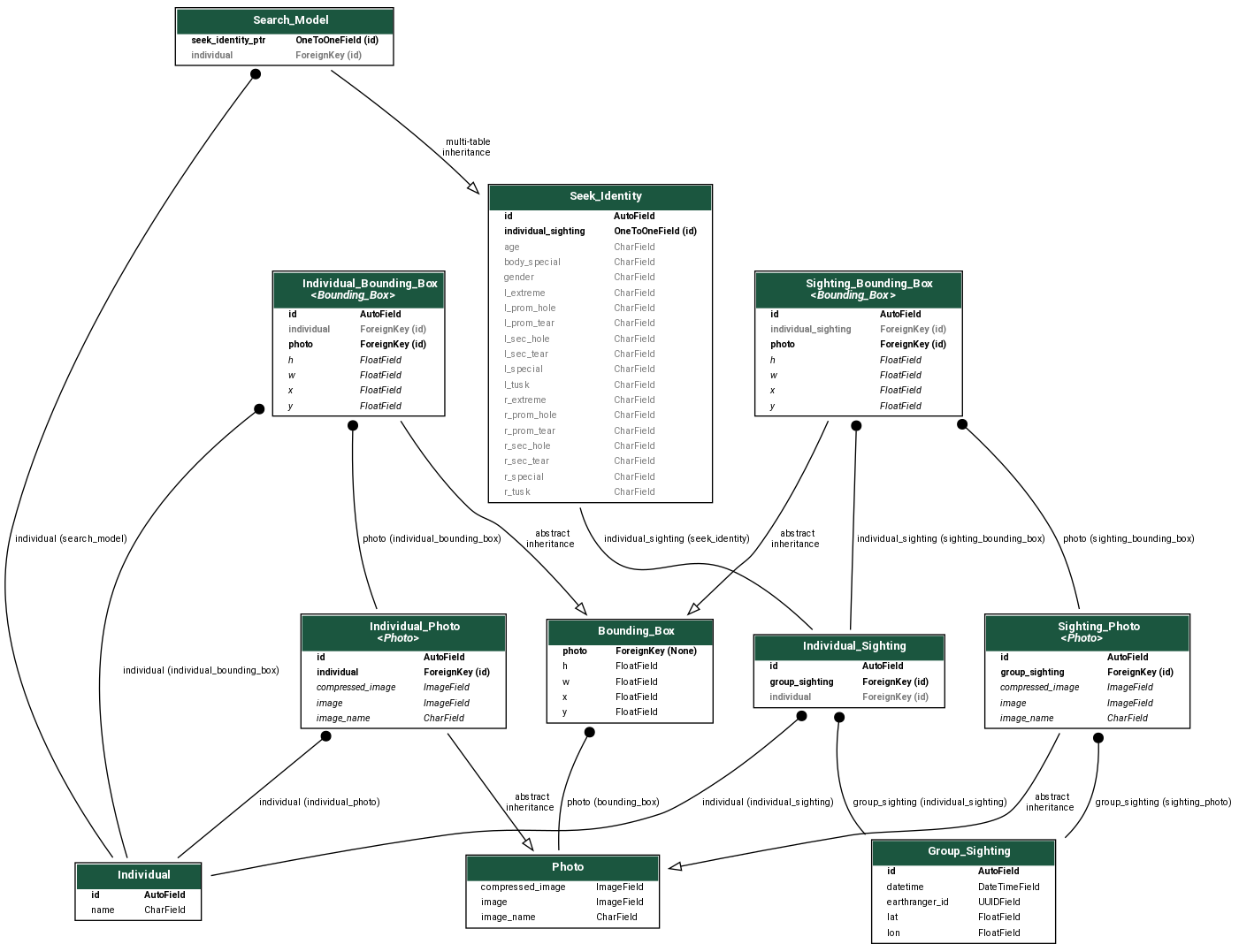}
\end{tabular}
\end{center}
\caption[detect-full] 
{ \label{fig:model} 
Database schema of major models in ElephantBook}
\end{figure}

\subsection{Human Expert Elephant Re-Identification}

Elephant re-identification is a difficult task, and ecologists may spend thousands of hours over their careers cataloging and charting features that can be used to distinguish elephants. These approaches are often heavily subjective and based on the interpretation and skill of the observer, making the process difficult to replicate across multiple observers or elephant populations. Quantitative approaches are needed to reliably re-identify elephants without dependence on the one or two experts typically available within an organization. One of the most successful existing methods of differentiating between elephants relies on comparison of the elephants' ears including notches, tears, holes, and other identifiable patterns. Several organizations (e.g., Save the Elephants, Elephants Alive, Elephant Voices, Amboseli Elephant Trust) use this expert-based approach for elephant identification.

Elephants Alive developed SEEK \cite{bedetti2020system}, which involves a comprehensive identification dataset comprised of photos, drawings, and codes of elephant ear patterns that were collected over 25 years (since 1996). The identification system has been refined over time to exclude observer bias and accelerate the photographic identification process. We believe SEEK is the least subjective or expert-reliant elephant re-identification system in-use by any organization to date. 

\subsection{Automated Animal Re-Identification}
The most commonly studied re-identification problems in computer vision focus on humans, with popular benchmarks and vast literature for human facial re-identification \cite{yi2014deep, liao2015person, xiong2014person}. There have been many recent successes in computer vision for automated species identification, in both camera trap data \cite{norouzzadeh2019deep, schneider2018deep, tabak2020improving, beery2018iwildcam, beery2019iwildcam,beery2020iwildcam, beery2019efficient, beery2020context, beery2020synthetic} and human-captured community science data \cite{van2015building, van2018inaturalist, van2021benchmarking, cui2019class, cui2018large, mac2019presence}. Automated re-identification of \textit{individual animals} using computer vision is an increasingly popular topic, with publications and workshops on the subject at major computer vision conferences \cite{WACVReID}. There are several excellent reviews of computer vision for animal re-ID \cite{schneider2019past,vidal2021perspectives, ravoor2020deep}. 

One of the main, and significant, differences between animal re-identification and other fine-grained categorization tasks is that populations are not fixed, making re-ID an open-set categorization problem \cite{zhou2018brief}. You must be able to recognize if and when an individual does not already exist in your database. The set of individuals might also be quite large: even for the relatively small global population of Grevy's zebra, your full set of identities would be ~8,000 individuals \cite{ontitastate, parham2017animal}.

The earliest proposed semi-automated re-identification systems go back as far as 1990, with works on whale re-identification based on human-annotated attribute similarity \cite{mizroch1990computer}. The next big breakthroughs in the field relied on traditional feature-engineered computer vision techniques for pattern matching (including SIFT-based feature matching) \cite{hiby1990computer, kelly2001computer, arzoumanian2005astronomical, burghardt2007fully, miele2020revisiting, hiby2009tiger, town2013m, berger-wolf_wildbook_2017, loos2013automated} and numerical representations of unique contours \cite{huele1998identification, hillman2003computer, ardovini_identifying_2008}. Animal re-identification, like most of computer vision, has seen significant advances with the onset of deep learning, including several neural-network-based approaches \cite{carter2014automated,freytag2016chimpanzee, brust2017towards, korschens_towards_2018, korschens2019elpephants, he2019distinguishing, li2018cow, Schofield_2019, Cheema_2017}. The field has recently explored metric-learning-based methods \cite{schneider2020similarity, zhou2018brief, deb2018face}, inspired by the success of these methods for human re-identification \cite{yi2014deep, liao2015person, xiong2014person}. Metric-learning methods are also more robust to open-set categorization, as they are similarity-based and require only a single example of an individual with which to compare, as opposed to the tens or hundreds of examples needed by data-hungry CNNs. Another common tactic for handling the open-set and data-scarce nature of re-identification is hybridizing deep networks for notable part localization with previous pattern or contour feature-based matching methods which do not require large amounts of training data per individual \cite{shukla2019hybrid, weideman_extracting_2020}.

\subsection{Automated Elephant Re-Identification}
\begin{figure}
\centering
\begin{subfigure}{.5\textwidth}
  \centering
  \includegraphics[height=3.4cm]{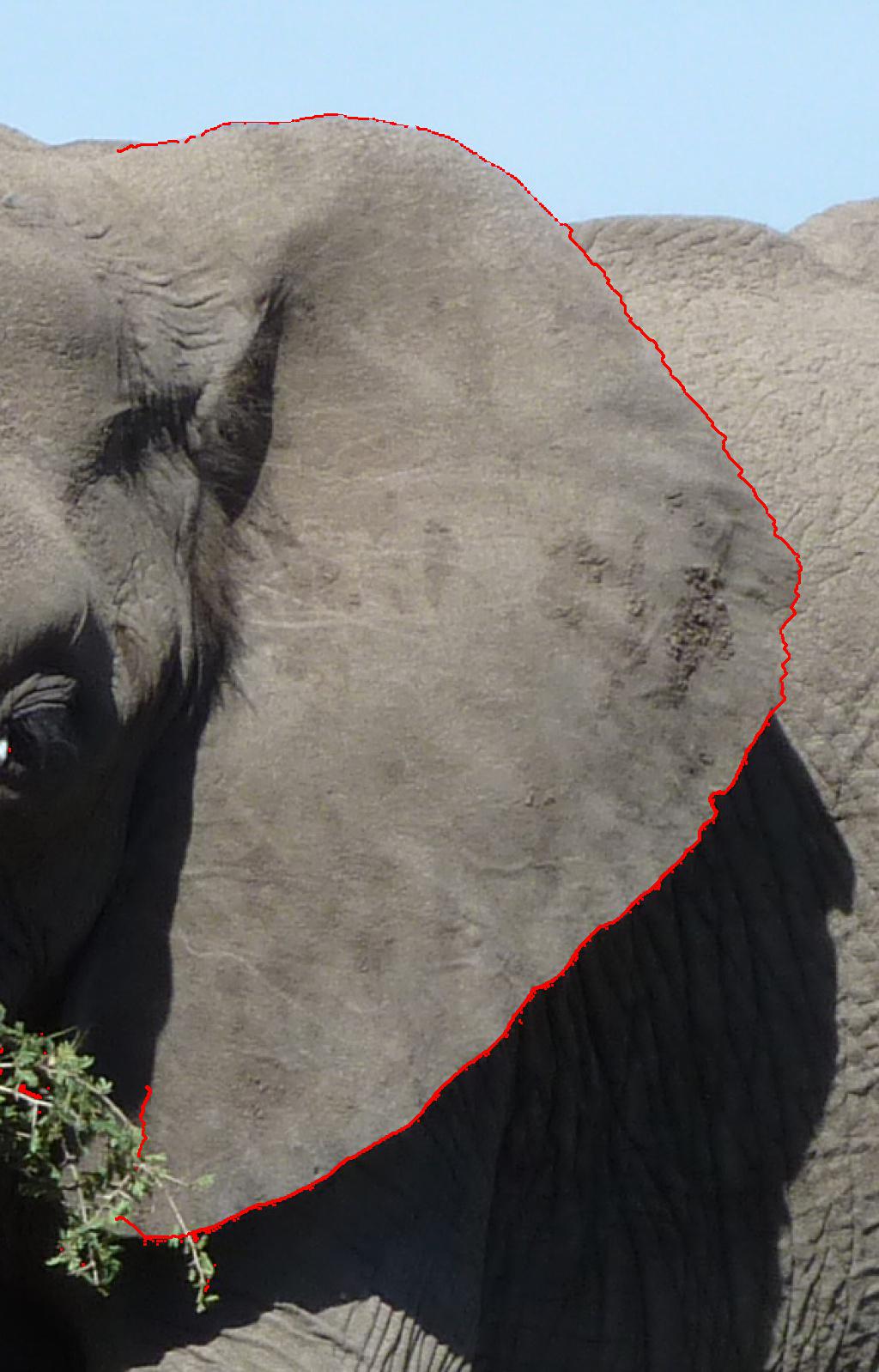}
  \includegraphics[height=3.4cm]{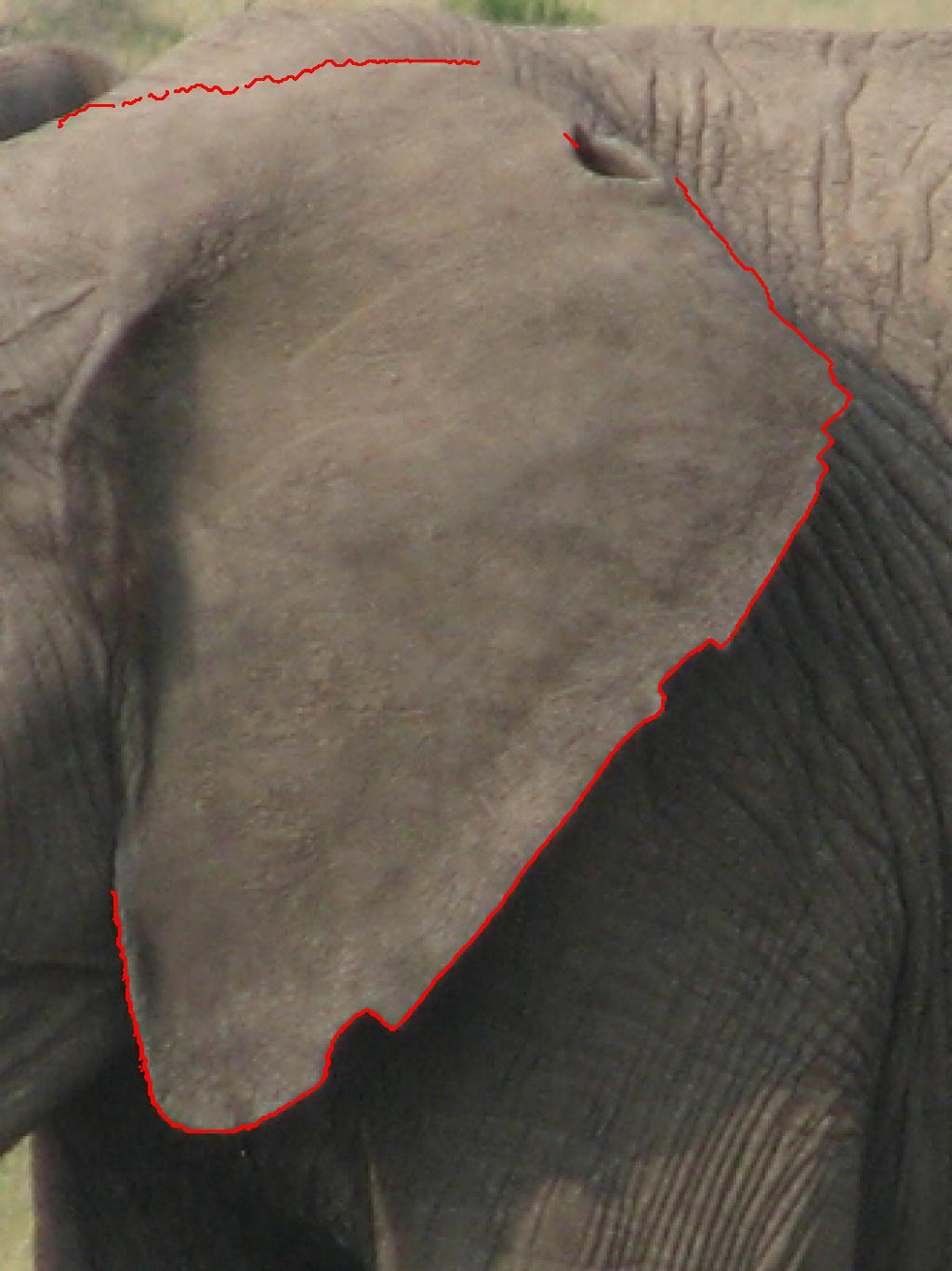}
  \includegraphics[height=3.4cm]{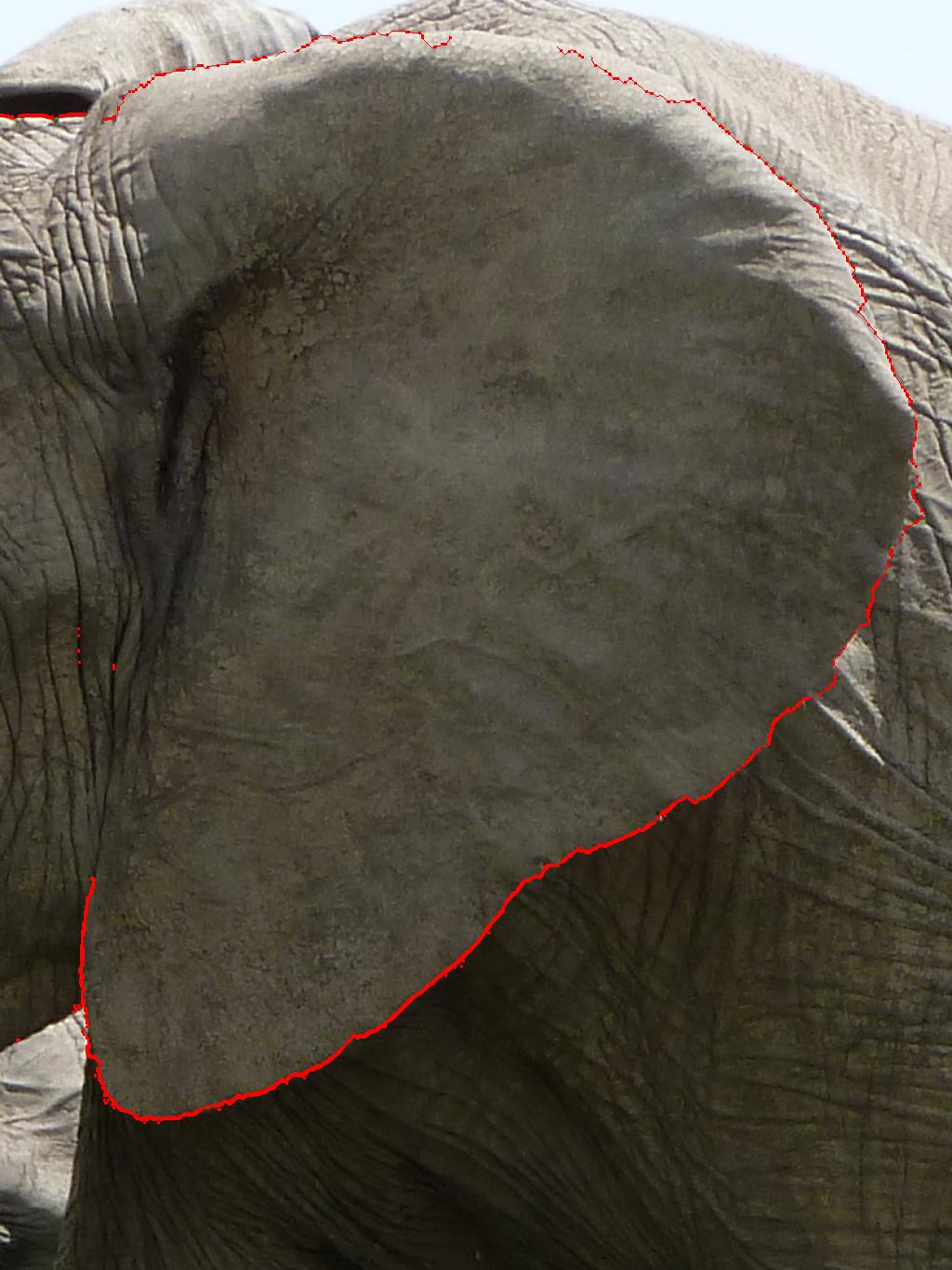}
  \caption{Successful curve extraction}
  \label{fig:sub1}
\end{subfigure}%
\begin{subfigure}{.5\textwidth}
  \centering
  \includegraphics[height=3.4cm]{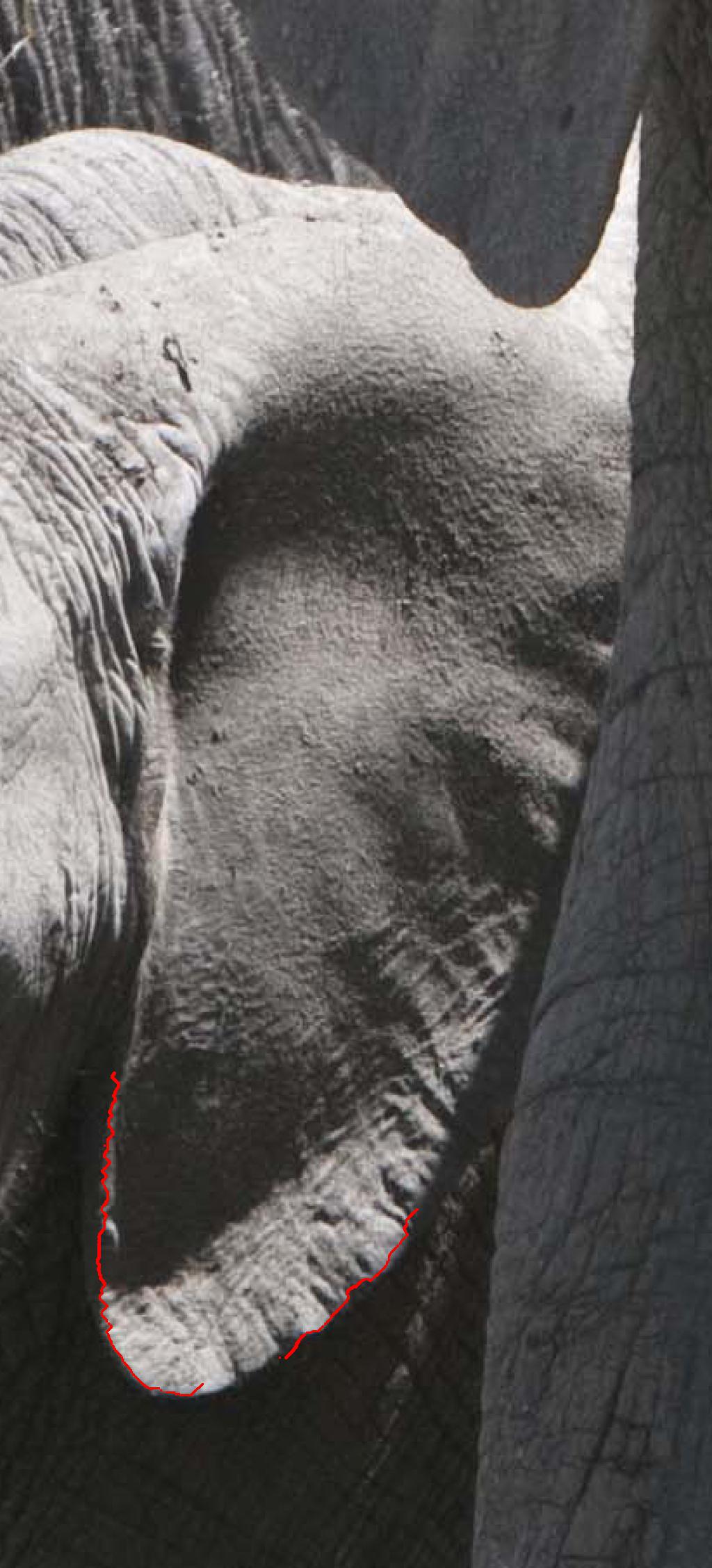}
  \includegraphics[height=3.4cm]{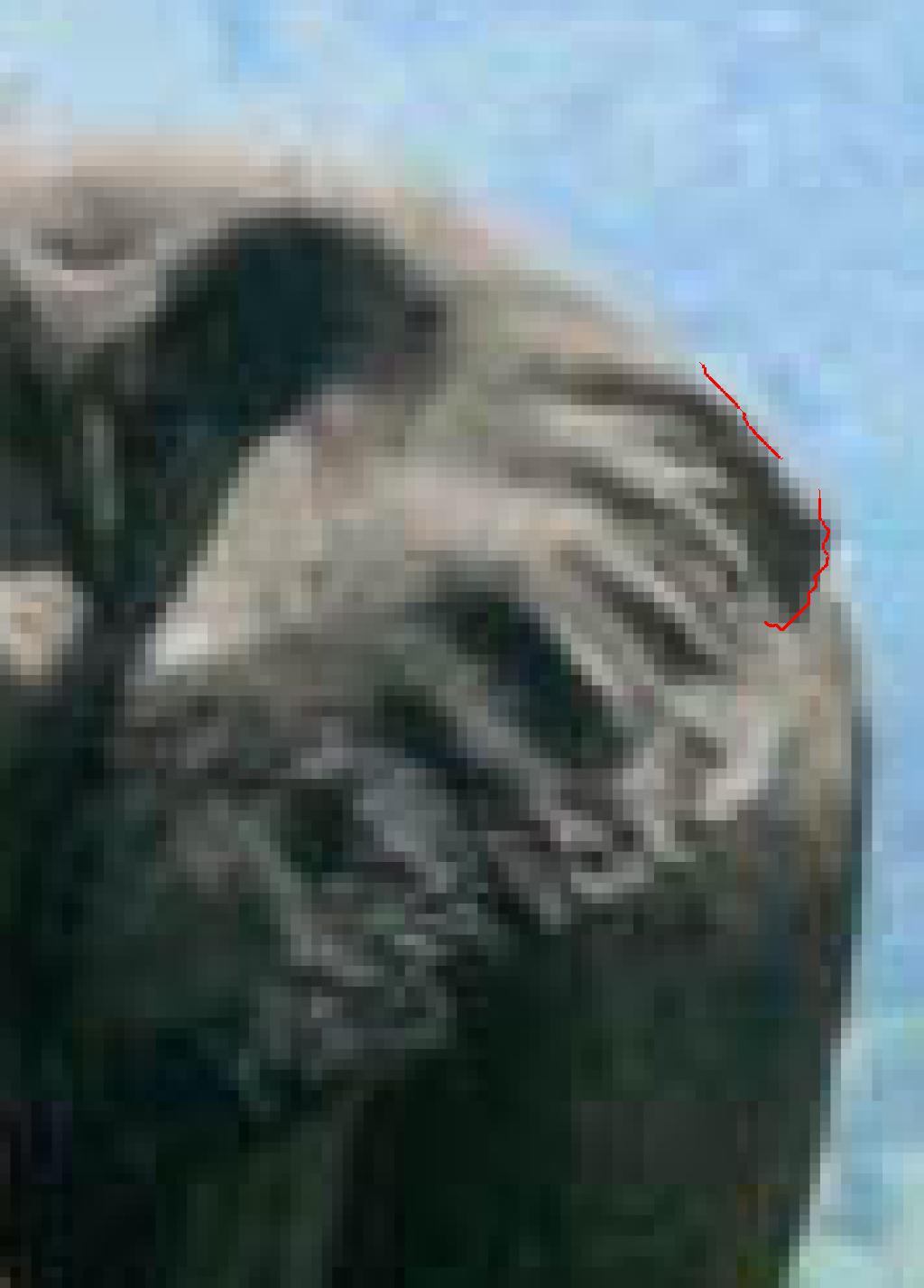}
  \includegraphics[height=3.4cm]{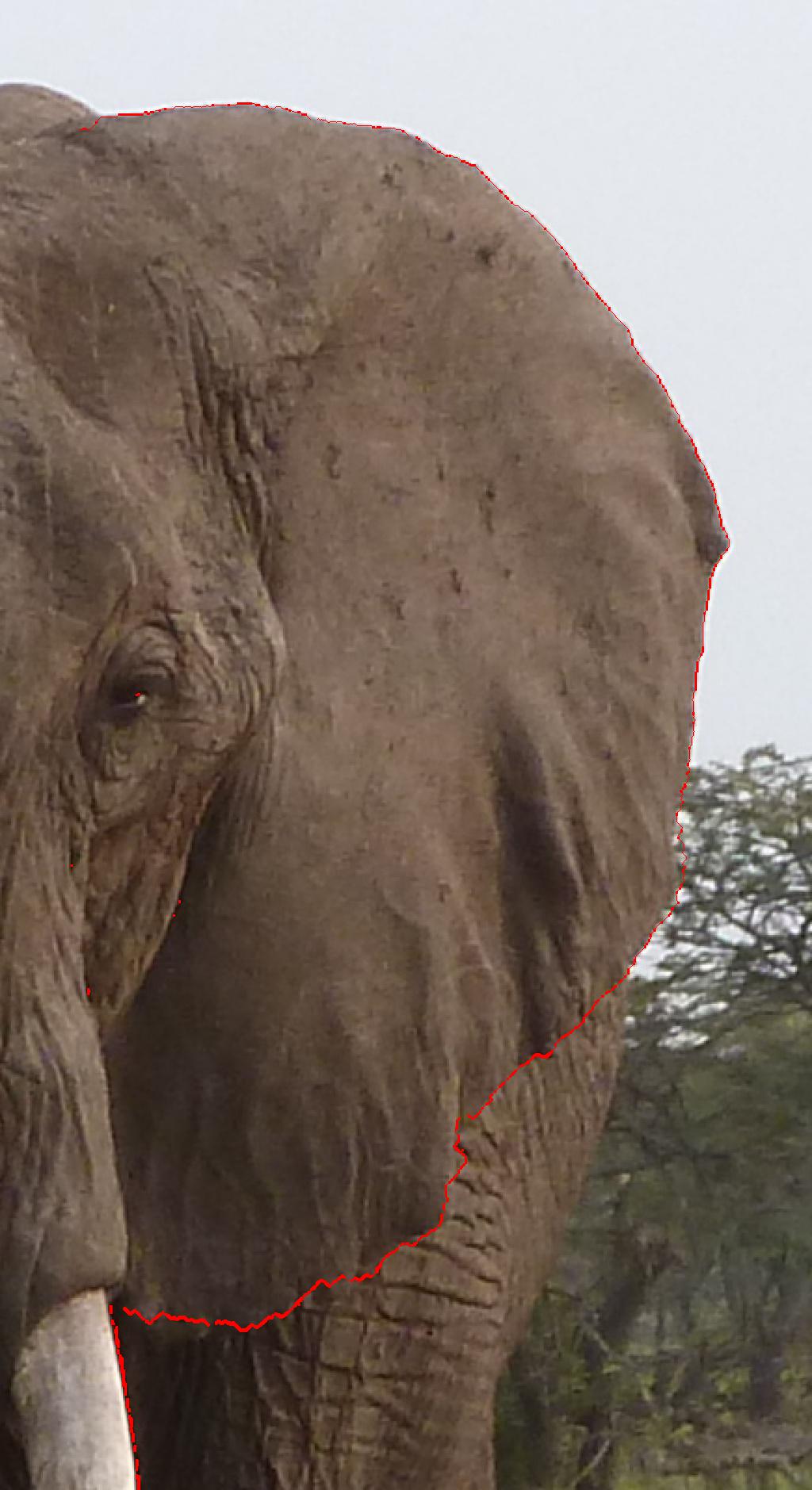}
  \caption{Failed curve extraction}
  \label{fig:sub2}
\end{subfigure}
\caption{Here we show successful and failed CurvRank examples. CurvRank is highly successful in high-quality, high-resultion imagery, but performance drops off in lower-resolution or blurry data as the edge of the ear is harder to distinguish.}
\label{fig:curve}
\end{figure}

In 2010, Dabarera and Rodrigo proposed an image-based algorithm to identify individual elephants based on full-frontal facial images \cite{dabarera_vision_2010}.
Korschens et al. proposed a matching algorithm based on human-labeled whole-head annotations, including the elephant’s ears and tusks, where present
\cite{korschens_towards_2018}. In an extension, they released a dataset, ELPephants, and demonstrated good results on a closed set of individuals with localized feature extraction using deep nets and SVM-based feature discrimination \cite{korschens2019elpephants}. Recent methods of robustly differentiating between elephant images in an open-set population rely on finding and matching the contours of the ear (Figure~\ref{fig:curve}), similar to many human-expert re-identification methods like SEEK. Multi-curve matching algorithms based on  human-annotated contours of elephant ears were proposed by Ardovini et al. \cite{ardovini_identifying_2008} and Weideman et al. \cite{weideman_integral_2017}. Weideman's CurvRank algorithm was originally designed for re-identification of whale flukes and dorsal fins. Recently, Weideman et al. \cite{weideman_extracting_2020} proposed an extension of CurvRank that is capable of automatically extracting matchable contours from images, and report strong results matching contours of elephant ears.

\section{ElephantBook}

Our solution, which we call ElephantBook, by default integrates with EarthRanger through its REST API to consume Group Sightings recorded by field teams. ElephantBook can also be reconfigured for use without EarthRanger if needed. It is web-based and built primarily with the Django Python package \cite{djangoproject}. This configuration allows our system to be both lightweight and easily reconfigurable.

\subsection{Human-in-the-Loop Re-Identification Pipeline}
\begin{figure}[ht]
\begin{center}
\begin{tabular}{c} 
\includegraphics[height=17.5cm]{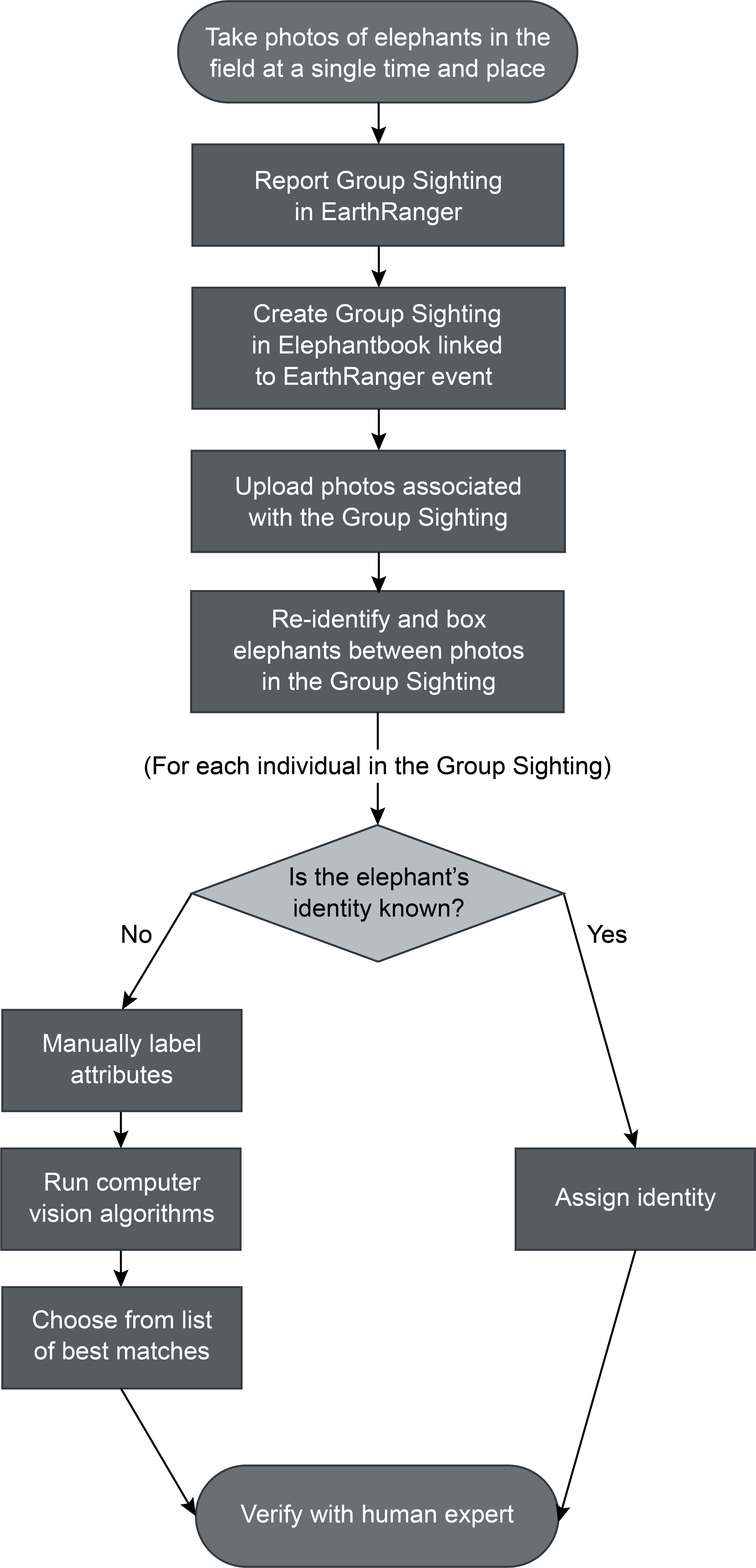}
\end{tabular}
\end{center}
\caption[detect-full] 
{ \label{fig:workflow} 
Workflow of elephant re-identification}
\end{figure}

\setcounter{subsubsection}{-1}
\subsubsection{Data Collection in the Field}
Rangers at the Mara Elephant Project routinely survey the Maasai Mara in search of elephants. Rangers record the time and location of every elephant sighting and submit it to EarthRanger. When possible, rangers photograph each elephant from multiple angles. If no photographs are taken, the event is resolved in EarthRanger, and no re-identification occurs.
\subsubsection{Adding a Group Sighting to ElephantBook via EarthRanger}
ElephantBook pulls a list of active elephant sighting events from EarthRanger. Users select the appropriate EarthRanger event and create a corresponding ElephantBook ``Group Sighting.'' A Group Sighting is one or more elephants spotted at the same time and place. This step is usually performed after returning from the field.
\subsubsection{Uploading Photos}
All photos taken at the same time and place of the Group Sighting are uploaded to ElephantBook. However, only photos labeled with boxes (in the next step) are used for re-identification.
\subsubsection{Boxing Elephants}
Once all photos are uploaded, elephants in each photo are boxed with an image annotation tool. While the human annotator likely will not know the name of each individual elephant in the Group Sighting photos, the annotator should be able to differentiate between elephants and identify the same elephant across multiple images. If it is impossible to tell elephants apart in a single instance, matching over a period of months is unlikely.
Boxes are marked with numbers unique to each specific elephant within the Group Sighting. This identification marking reduces the number of matches we must make from the sum of the number of elephants in all photos to the number of actual elephants sighted. See Section \ref{bounding_box_annotation} for more details.
\subsubsection{Human Attribute Labeling}
An ``Individual Sighting'' is created for each elephant identified in the previous step. An Individual Sighting is an elephant encounter at a single time and place, and it is always connected to a parent Group Sighting.
Manual attribute-labeling is performed for each Individual Sighting. We use the recently-developed SEEK coding system \cite{bedetti2020system}. See Section \ref{seek} for more details.
\subsubsection{Computer Vision}
Confidence-producing computer vision matching algorithms are run to identify potential matches. See Section \ref{computer_vision} for more details.
\subsubsection{Matching}
Manual attributes are combined with the output from the computer vision algorithms to provide a list of the most likely previously identified elephants. See Section \ref{matching} for more details.
\section{Bounding-Box Annotation} \label{bounding_box_annotation}
We customized an open-source online bounding-box annotation tool from the Visipedia project \cite{belongie_visipedia_2016}. Because annotators need to match individual elephants across photos taken at a sighting, a second pane was added to allow annotators to compare multiple photos at once.

\begin{figure}[ht]
\begin{center}
\begin{tabular}{c} 
\includegraphics[height=5cm]{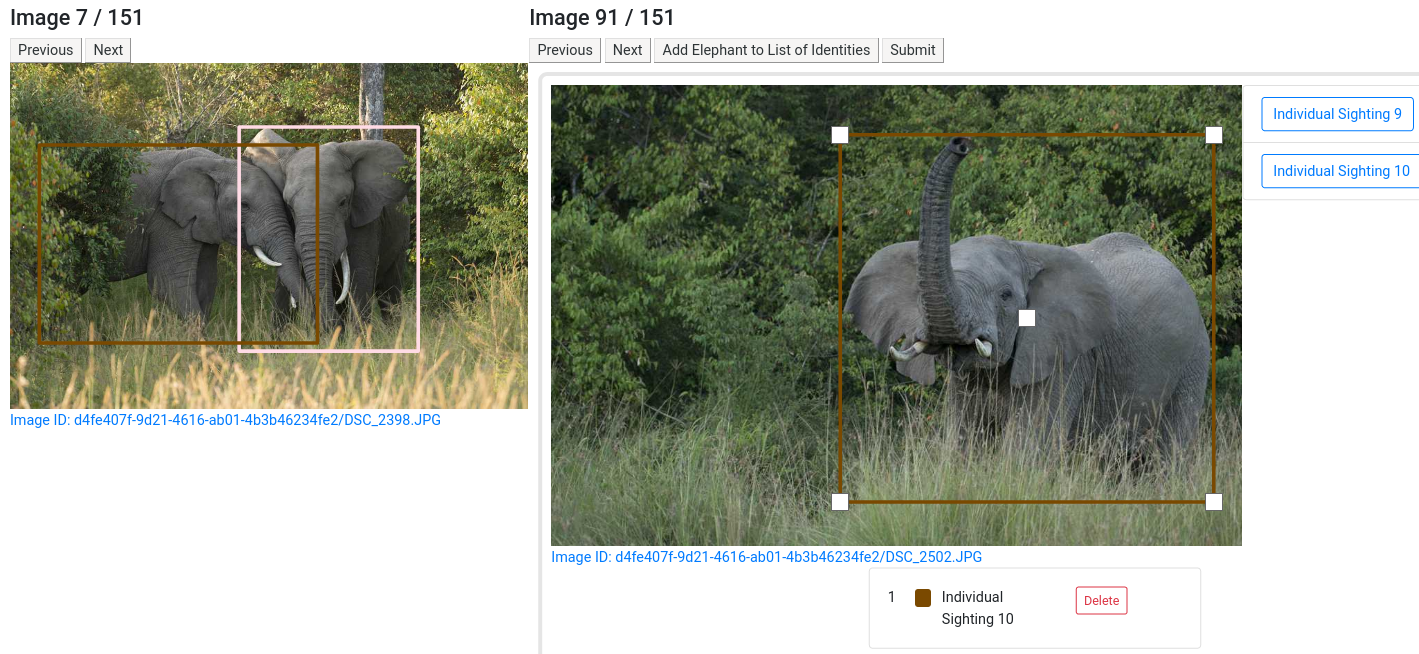}
\end{tabular}
\end{center}
\caption[detect-full] 
{ \label{fig:annotator} 
Bounding-box annotation interface}
\end{figure}

\section{SEEK} \label{seek}

\begin{figure}[ht]
\begin{center}
\includegraphics[page=1,width=0.805\textwidth]{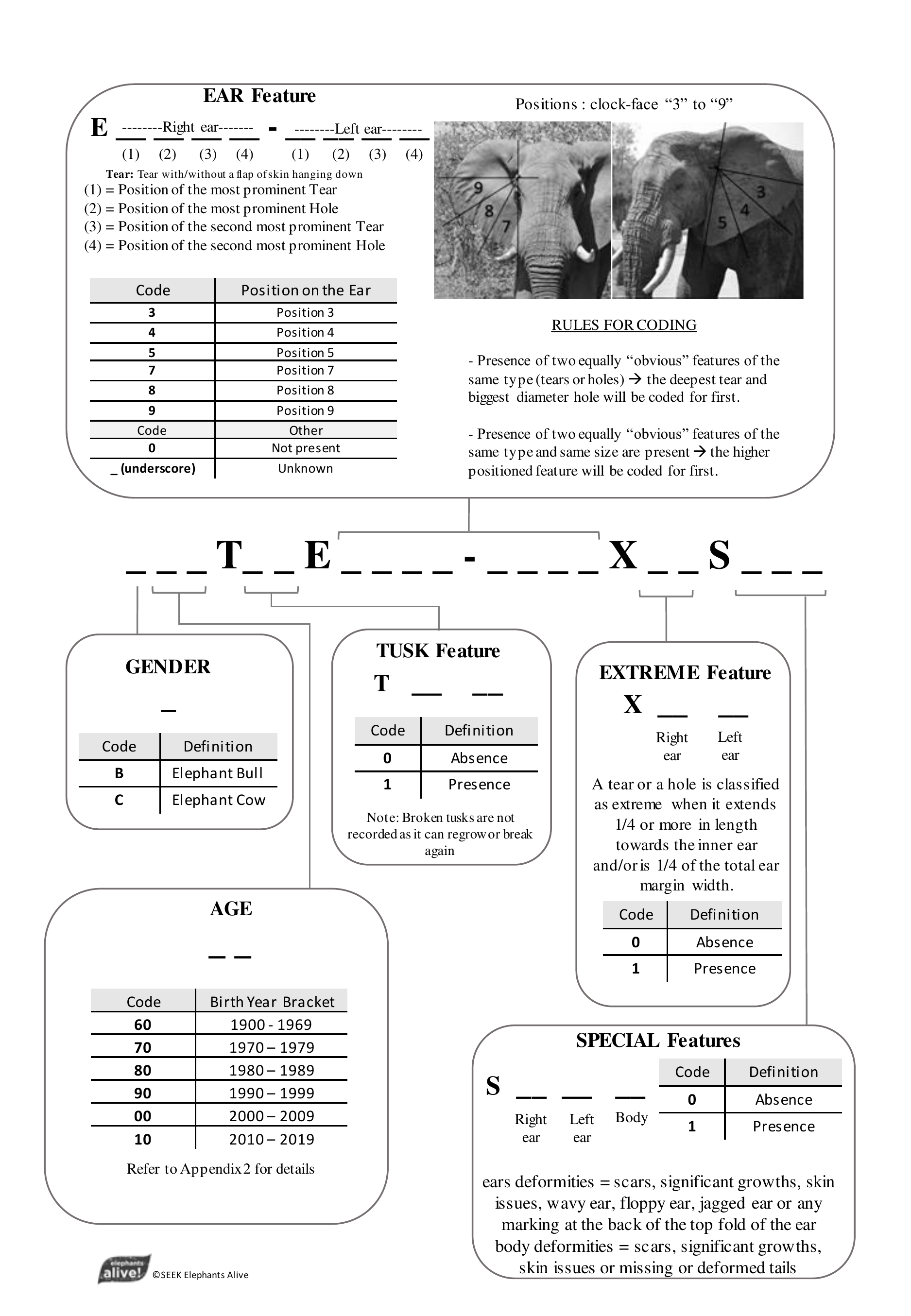}
\end{center}
\caption[detect-full] 
{ \label{fig:seek_system} 
The SEEK coding system}
\end{figure}

In SEEK, each elephant is assigned a unique descriptive code which is used to narrow the set of potential matches that must be considered by human experts. The code begins with the elephant’s sex and age, followed by the presence or absence of tusks (Figure~\ref{fig:seek_system}). The code further defines the type and position of prominent and secondary tears and holes found on the right and left ears. Finally, it notes the presence of any extreme features on the ears and body, such as a missing tail.

\begin{figure}[ht]
\begin{center}
\begin{tabular}{c}
\includegraphics[height=10cm]{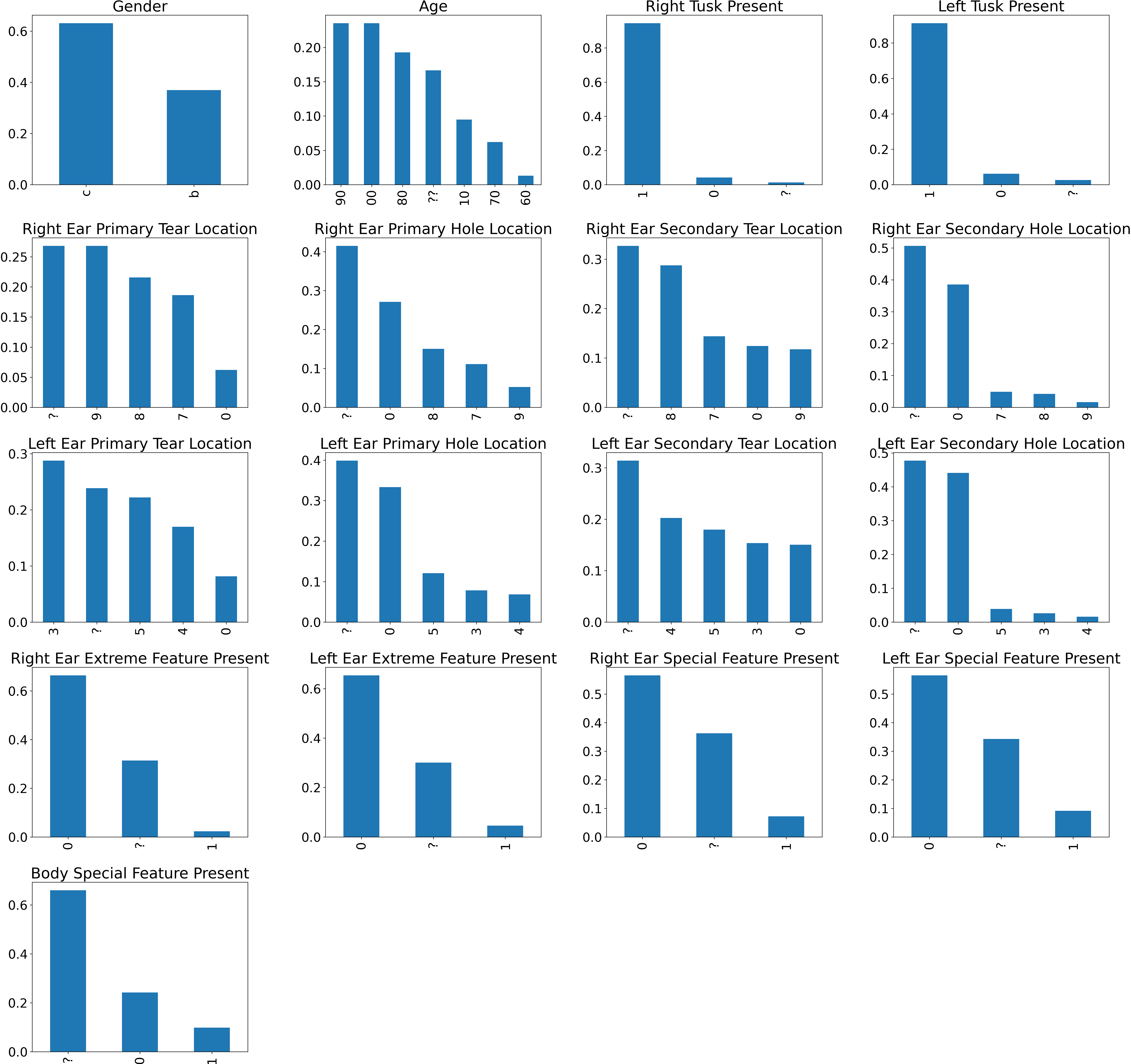}
\end{tabular}
\end{center}
\caption[detect-full] 
{ \label{fig:seek_frequency} 
Frequency of SEEK attributes}
\end{figure}

\begin{figure}[ht]
\begin{center}
\begin{tabular}{c}
\includegraphics[height=10cm]{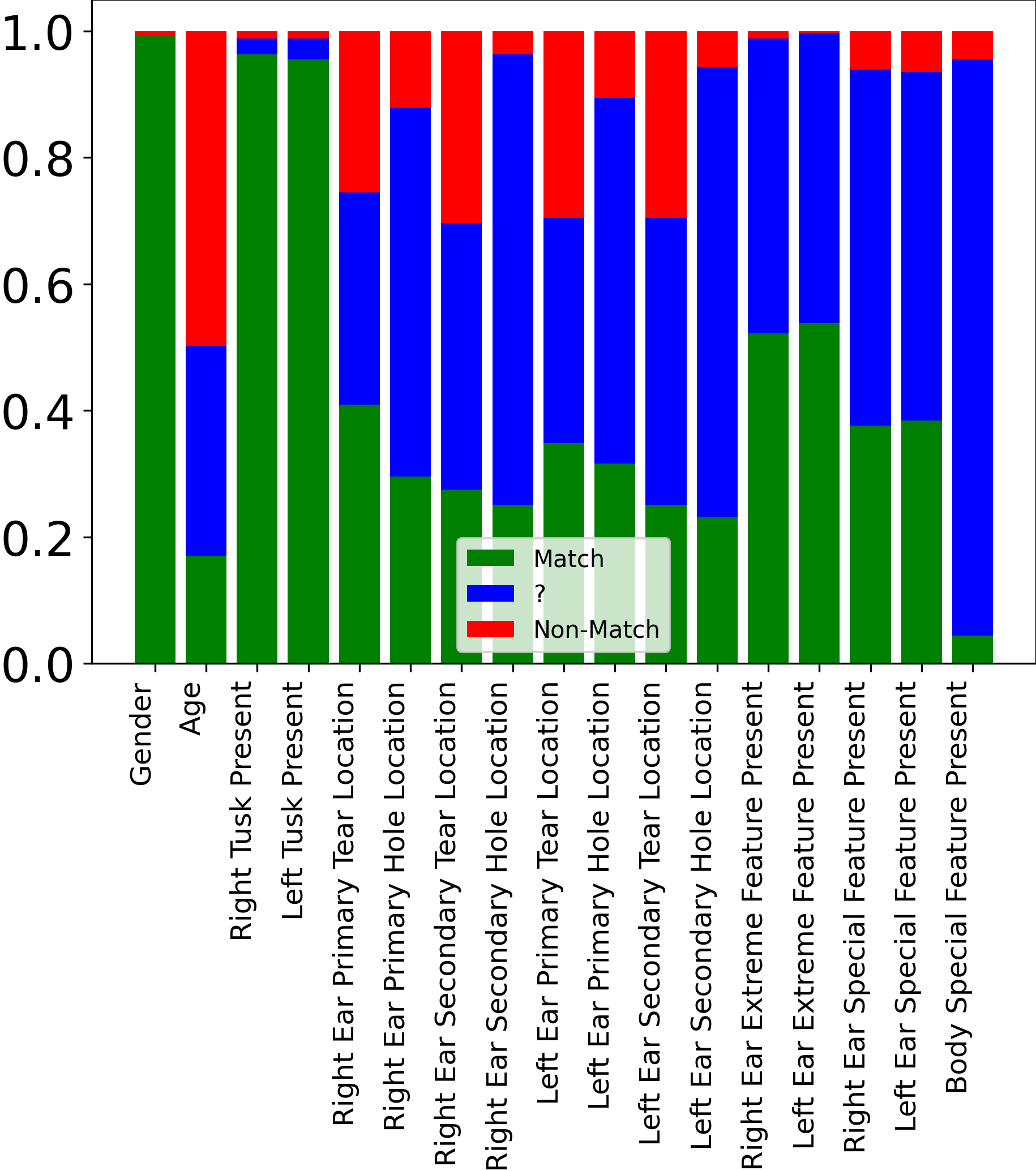}
\end{tabular}
\end{center}
\caption[detect-full] 
{ \label{fig:seek_agreement_for_pairs} 
Agreement of an\\
notators by attribute for pairs of SEEK codes on the same individual}
\end{figure}

\section{Computer Vision} \label{computer_vision}
\subsection{Elephant Ear Localization}
To allow CurvRank to focus on the ear, the localization of which is key to extracting accurate ear contours, we trained a simple elephant-ear detector.

\begin{figure}[ht]
\begin{minipage}{0.49\linewidth}
\begin{center}
\begin{tabular}{c} 
\includegraphics[height=5cm]{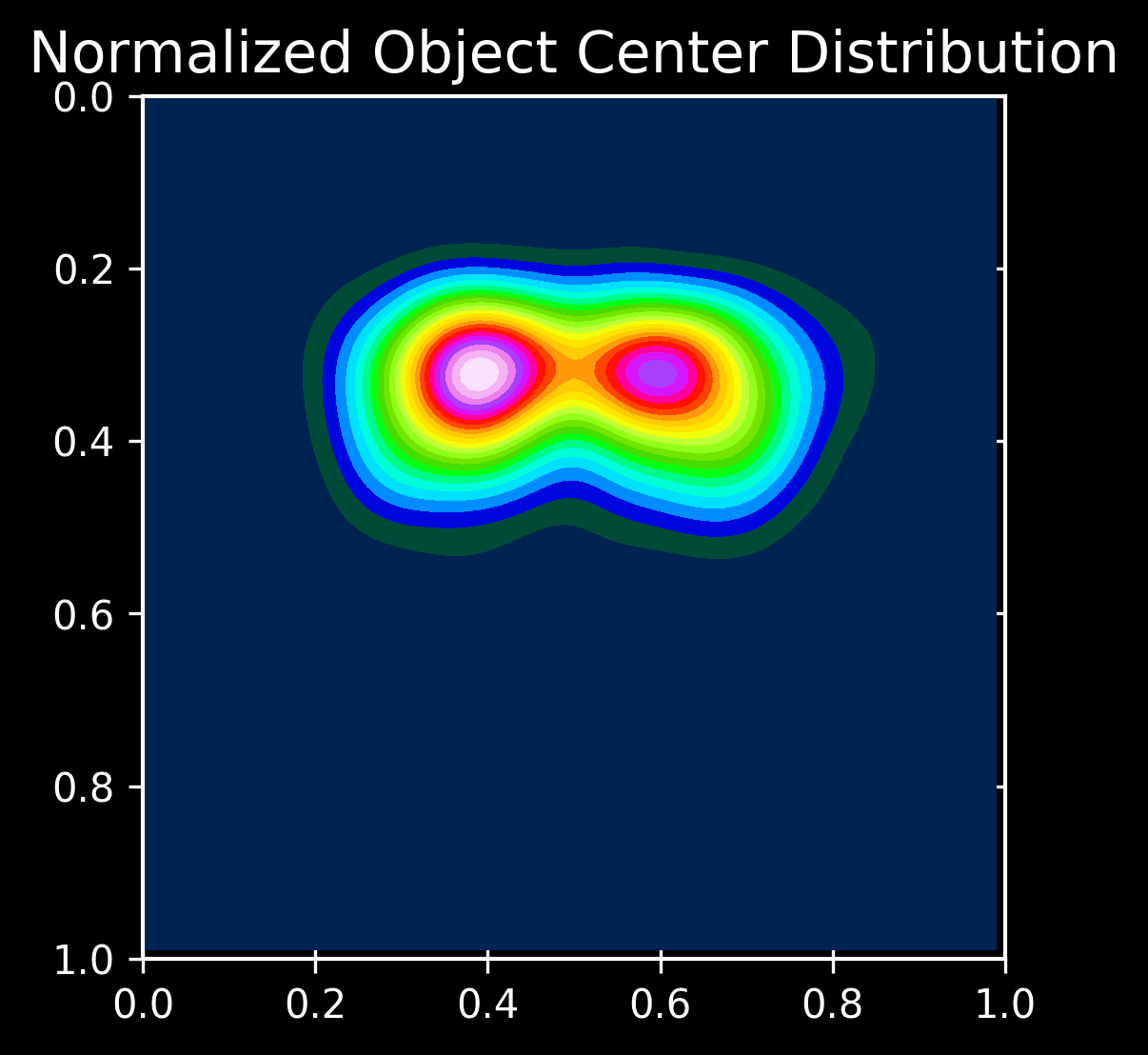}
\end{tabular}
\end{center}
\caption[detect-full] 
{ \label{fig:center_distribution} 
We visualize the center of the ground truth annotated boxes across our training set, and see that there is a strong bias in the imagery towards ears being in the upper center of the image, with modes slightly to the right and left}
\end{minipage}
\end{figure}

The ELPephants dataset from the Elephant Listening Project \cite{korschens2019elpephants} -- consisting of images of African forest elephants visiting the Dzanga bai clearing in the Dzanga-Ndoki National Park in the Central African Republic -- was used for training and validation. After removing duplicates, the dataset consisted of 1935 images of 276 unique individuals. The dataset is provided with the identities of the elephants, but each image was manually annotated for bounding-boxes of left and right ears. Only ears where the contours were fully visible were annotated. Annotations were made on 910 left ears and 1,045 right ears (Figure~\ref{fig:center_distribution}). Two-hundred randomly sampled images were reserved for object detection validation.

We trained a Faster R-CNN object detection model \cite{ren2015faster} with a ResNet-50 backbone \cite{he2015deep} and added Feature Pyramid Networks (FPN) \cite{lin2017feature} in Pytorch \cite{paszke2019pytorch}. Beginning with a model checkpoint pretrained on the Microsoft COCO dataset \cite{lin2014microsoft}, we trained our model on 1,735 images to detect and categorize left and right ears. Our detector achieves a Mean Average Precision (mAP) \cite{lin2014microsoft} of 95\% on our held-out test dataset of 200 randomly-selected images.

\subsection{Matching Ear Contours with CurvRank}

After extracting ear images from out ear detection model, we use CurvRank \cite{weideman_extracting_2020} to filter possible matches.

CurvRank was initially developed to recognize individual cetaceans based on contours of flukes and dorsal fins  \cite{weideman_integral_2017}. As elephant ears are also a strong re-identifiable feature and are delineated by a contoured edge, applying CurvRank to elephant re-identification was an intuitive next step, and the authors determined the transferability of the matching algorithm from cetaceans to elephants by analyzing results on hand-drawn contours.

Recently the CurvRank authors proposed a deep-learning based algorithm to automatically extract the contours used as input to their matching algorithm \cite{weideman_extracting_2020}. They evaluated this automated approach on humpback whales and African savanna elephants, with impressive results. Their method relies on two fully convolutional neural networks for curve extraction, one coarse-grained and one fine-grained  (CG-FCNN and FG-FCNN). Annotators initially traced the identifying contour in cropped images with a broad line, using a single brushstroke, to produce coarse-grained training data for the CG-FCNN. In the second step, the FG-FCNN is trained to predict for each pixel in the (ear or fluke) image the probability that it would be covered by the coarse brush stroke, producing a probability image at the same resolution as the initial image. By using the coarse, easily extracted training data to train the FG-FCNN, tedious manual effort is avoided. These pixel-level probability maps guide the third step: a shortest path contour extraction algorithm. 
 
Once the contour is extracted, it is represented as an ordered sequence of (x, y) coordinate pairs. Then CurvRank builds an integral curvature by sliding multiple disks of increasing radius
along the contour \cite{weideman_extracting_2020}. For each scale, every point is represented as the ratio of the areas of the disk for that scale on either side of the contour \cite{weideman_extracting_2020}. Feature keypoints
are defined at local extrema of the integral curvature representation \cite{weideman_extracting_2020, hughes2017automated}, and feature descriptors are extracted from the regions between all pairs of keypoints. This set of feature descriptors forms a densely sampled, overlapping representation of the entire individual contour across multiple scales. Match similarity is determined and possible matching individuals ranked via the local naive Bayes nearest neighbors (LNBNN) algorithm \cite{mccann2012local}.

The method reported a top-1 matching accuracy of 84\% for high-quality, high-resolution images of elephant ears on the closed set of 132 individuals on which the model is also trained \cite{weideman_extracting_2020}. The authors remark that elephant ear recognition is more difficult than whale fluke identification, due to challenges of contour extraction against more-highly textured image backgrounds and because the identifying information is more localized and subtle.

\section{Matching} \label{matching}
To allow rangers to efficiently identify an individual from the large set of previously encountered elephants, rangers are presented with a ranked list of possible matches to visually examine. These matches are computed with a score function that is a linear combination of manual attribute differences and computer vision matching confidence. Each SEEK attribute of the new Individual Sighting is compared to the SEEK code of all known Individuals. For each attribute in the codes, the distance is zero if the attributes match, one if they differ, and 0.6 if either of them contain a wildcard character. Additionally, the weight of the age component of the distance is set to 0.4 because of the known difficulty in accurately aging elephants (Figure \ref{fig:seek_agreement_for_pairs}). The mean of these differences is taken. The weighting parameters were learned separately on a training set of codes to optimize matching accuracy.

CurvRank produces an unbounded matching score between the new Individual Sighting and all Individuals. A greater score indicates greater likelihood of a true match. CurvRank scores are subtracted from the SEEK score and multiplied by 0.1. The constant 0.1 was learned in a training set of SEEK codes and CurvRank contours.

\subsection{Evaluating Matching Accuracy}
To evaluate the robustness of SEEK, CurvRank, and our proposed combination of the two, we trained a non-expert team of seven college undergraduates to perform SEEK labeling, and we collected labels from two to three students annotator for a set of Individual Sightings from the Elephant Voices collected by Joyce Poole \cite{joyce_data}. In total, we have three annotations for 75 Individual Sightings and two annotations for 26 Individual Sightings.

\begin{figure}
\centering
\begin{subfigure}{.5\textwidth}
  \centering
  \includegraphics[width=.8\linewidth]{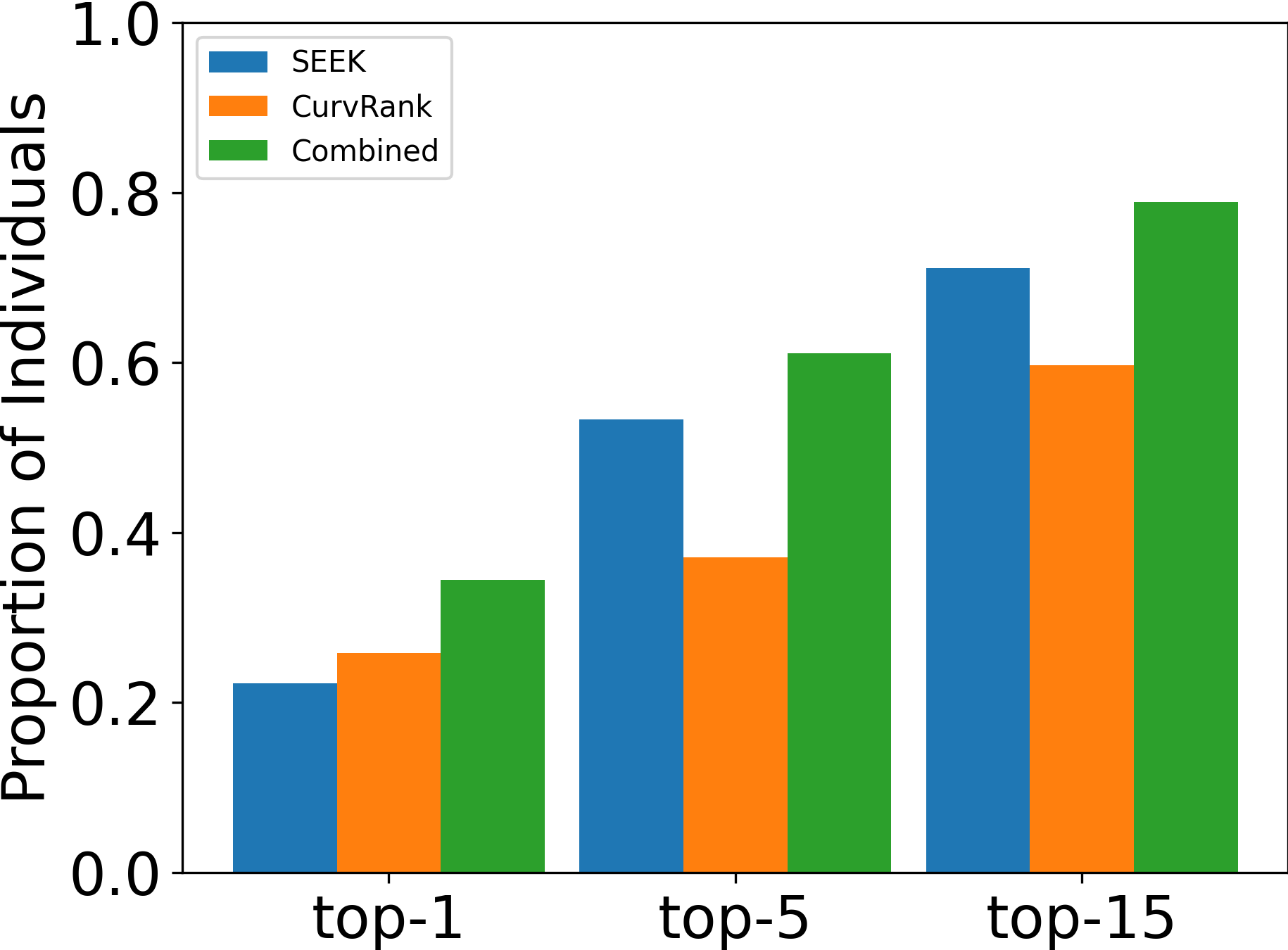}
  \caption{One code per individual}
  \label{fig:match1}
\end{subfigure}%
\begin{subfigure}{.5\textwidth}
  \centering
  \includegraphics[width=.8\linewidth]{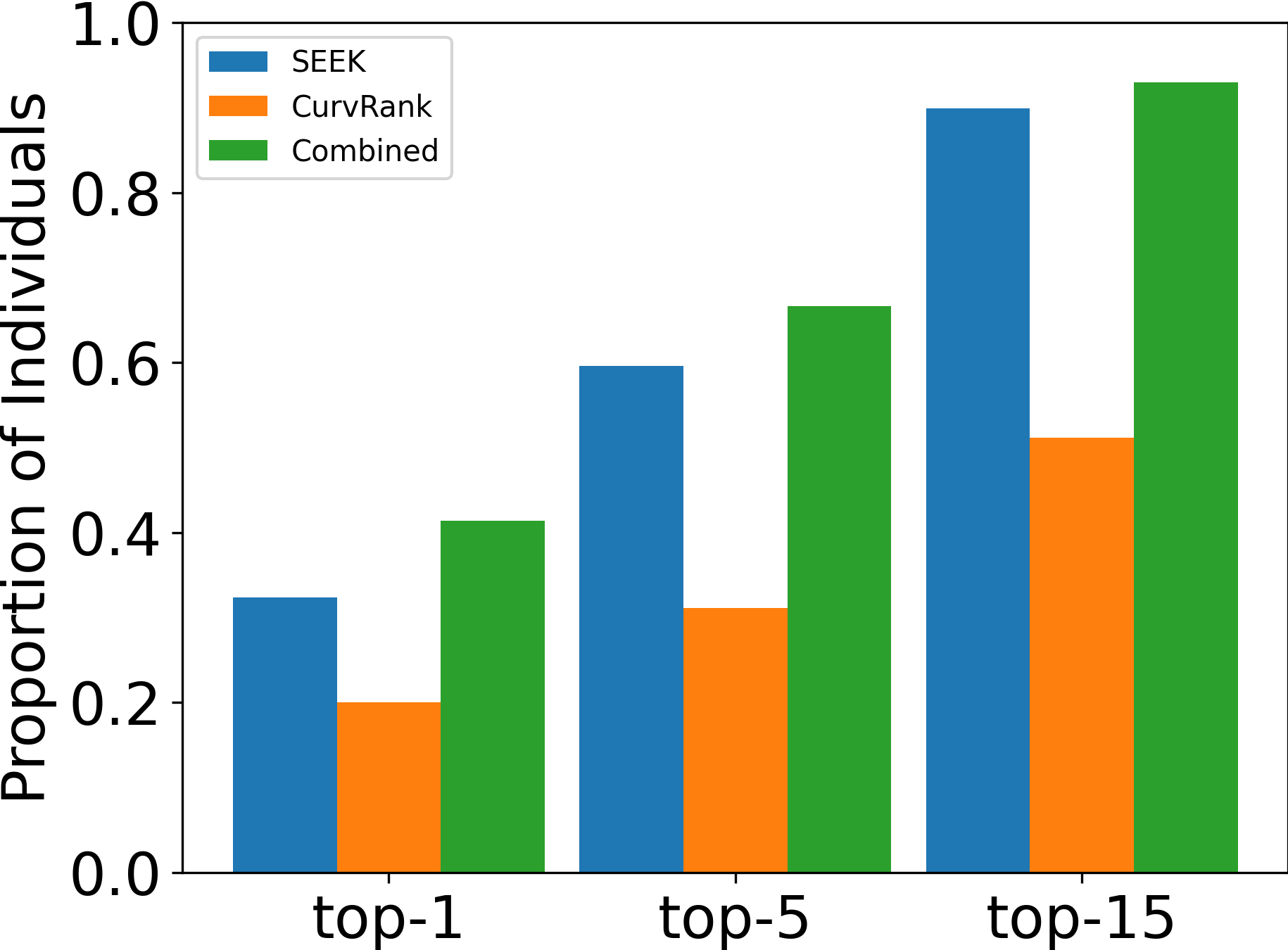}
  \caption{Two codes per individual}
  \label{fig:sub2}
\end{subfigure}
\caption{Comparing matching accuracy for our SEEK-based matching algorithm, CurvRank, and our hybrid SEEK-CurvRank aggregated approach. We see that SEEK and CurvRank are complementary, with the combined approach outperforming either method on its own for tests with both one and two database example for each individual.}
\label{fig:seek_agreement_by_num}
\end{figure}

We held out individuals that had at least two SEEK code annotations and two right-ear CurvRank contours. There are 45 individuals with a pair of SEEK codes and 33 individuals with a triplet of codes. Comparisons of top-k matching performance for SEEK, CurvRank, and our ensembled approach can be seen in Figure \ref{fig:seek_agreement_by_num}. We observe that matching performed with SEEK codes generally outperforms that of CurvRank alone, but that a combined approach is able to leverage the best of both, leading to more accurate matching. Using our combined system, with only two previous sightings of an individual in our database, we are able to match to the correct individual within the top 15 for 92.9\% of sightings, and within the top 5 for 66.7\%, helping rangers reduce the time needed to find the correct matched individual in the database. As Mara Elephant Project continues to collect and label Individual Sightings we will continue to analyze and hopefully improve matching performance. We expect additional sightings to improve accuracy, as it presents more potential sightings per individual to match with correctly. However, as we collect additional sightings we will also be increasing the number of individuals in the database, making the matching task more nuanced and potentially more challenging.

\section{Mara Elephant Project Initial Deployment}
The Mara Elephant Project began using ElephantBook in January 2021 after a six-month prototyping period and so far has logged 140 Group Sightings and 251 Individual Sightings and has ingested and boxed 10,462 images of elephants. Beginning in March 2021, the organization has hired and trained a full-time team of four research assistants for collecting elephant sightings in the field, processing photos, and developing SEEK codes for individual elephants. Initial training on both field methodology for cataloging elephant Group Sightings and in the use of ElephantBook and SEEK labeling took one week. MEP's goal is to characterize and document the majority of the Mara's ~2500 individuals. Extension of the ElephantBook system with partner organizations in Tanzania would further enable documenting the greater, connected elephant population stretching south into the Serengeti and consisting of >7000 individuals. 

Initial experience using ElephantBook is that it is an intuitive system that mimics a typical re-identification workflow. Optimizations for low-bandwidth connections, such as compression of photos before viewing them, but also keeping original full-resolution versions available for detailed scrutiny by a SEEK coder, have greatly improved the user experience. Boxing individuals has been relatively straight-forward even for novice users. Accurately labeling SEEK codes is perhaps the most challenging component of the ElephantBook system, particularly the correct estimation of age category which requires considerable expertise and, to a lesser degree, the determination of sex. 

\section{Conclusion and Future Work}
We have built a robust semi-automated system for human-in-the-loop elephant re-identification, and we have deployed our system on the ground in the Greater Mara Ecosystem. This system allows the Mara Elephant Project to track a much larger population of elephants over time, as they will no longer need to collar an elephant to track its movements. The system is a needed tool to assist in their vital elephant conservation efforts. As we move forward, we will expand ElephantBook to additional parks, including the Grumeti Game Reserve in Serengeti National Park in Tanzania and Greater Limpopo Transfrontier Conservation Area in South Africa and Southern Mozambique.

In the coming months, we will continue to collect new elephant sightings and refine our matching system to further reduce the human effort needed for re-identification. We plan to investigate automating SEEK coding and integrating additional computer vision methodology into our system, building learned representations of individual elephants beyond their ear contours. The data collected will also allow us to further analyze how these elephant features change over time and allow us to conduct deeper analysis of our current system on an expanding set of known elephants.

\begin{acks}
We would like to thank the entire team at the Mara Elephant Project for their efforts in deploying this system. This work was supported, through funding, data storage, and computing resources, by Microsoft AI for Earth, the Caltech Resnick Sustainability Institute, and NSFGRFP Grant No. 1745301, the views are those of the authors and do not necessarily reflect the views of these organizations.
\end{acks}

\bibliographystyle{ACM-Reference-Format}
\bibliography{main}

\end{document}